\pdfoutput=1

\documentclass[11pt, leqno]{article}

\usepackage{ACL2023}

\usepackage{times}
\usepackage{latexsym}

\usepackage[T1]{fontenc}

\usepackage[utf8]{inputenc}

\usepackage{microtype}

\usepackage{inconsolata}

\usepackage{enumerate}
\usepackage{textgreek}

\usepackage{linguex}
\usepackage{multicol}
\usepackage{multirow}
\usepackage{booktabs}
\usepackage{graphicx}
\usepackage{float}
\usepackage{arydshln} 
\usepackage{nicematrix,tikz}
\usepackage{makecell}
\usepackage{subcaption}
\usepackage{rotating}

\title{Generalizable Sarcasm Detection Is Just Around The Corner, Of Course!}

\author{Hyewon Jang$^{1}$, Diego Frassinelli$^{1,2}$ \\
$^{1}$Department of Linguistics, University of Konstanz, Germany \\ $^{2}$Center for Information and Language Processing, LMU Munich, Germany \\
        \texttt{\{hye-won.jang, diego.frassinelli\}@uni-konstanz.de}\\
}

\begin{document}
\maketitle
\begin{abstract}

We tested the robustness of sarcasm detection models by examining their behavior when fine-tuned on four sarcasm datasets containing varying characteristics of sarcasm: label source (authors vs.\ third-party), domain (social media/online vs.\ offline conversations/dialogues), style (aggressive vs.\ humorous mocking). We tested their prediction performance on the same dataset (intra-dataset) and across different datasets (cross-dataset). For intra-dataset predictions, models consistently performed better when fine-tuned with third-party labels rather than with author labels. For cross-dataset predictions, most models failed to generalize well to the other datasets, implying that one type of dataset cannot represent all sorts of sarcasm with different styles and domains. Compared to the existing datasets, models fine-tuned on the new dataset we release in this work showed the highest generalizability to other datasets. 
With a manual inspection of the datasets and post-hoc analysis, we attributed the difficulty in generalization to the fact that sarcasm actually comes in different domains and styles. We argue that future sarcasm research should take the broad scope of sarcasm into account.

\end{abstract}

\section{Introduction}

Sarcasm can be used to hurt, criticize, or deride \cite{colston1997salting, frenda2022unbearable, keenan1999young, kreuz1989sarcastic} but also to be mocking, humorous, or to bond \cite{dews_why_1995, gibbs_irony_2000, pexman_does_2002}. It comes in many different shapes such as understatement, hyperbole, rhetorical questions \cite{leggitt_emotional_2000}, deliberate falsehood \cite{glucksberg1995commentary, riloff_sarcasm_2013}, or self-deprecation \cite{abulaish_self-deprecating_2018}. Given such complexity and diversity of sarcasm, one should ask whether current sarcasm detection models are robust enough to be easily applied to any kind of sarcasm. This question is particularly valuable in computational linguistics, as most prior work addressing sarcasm 
adopted a narrow working definition of sarcasm as `saying the opposite of the true message (often with the intent to be hurtful)' \cite{cai-etal-2019-multi, frenda2022unbearable, ghosh_magnets_2017, joshi_harnessing_2015, pan-etal-2020-modeling}. This definition is based on the traditional theory of sarcasm by \citet{brown1978universals}, but is not 
at all comprehensive 
\cite{oprea-etal-2021-chandler, sperber_wilson_1981}.

Prior work in sarcasm detection has used one dataset at a time that only includes a specific style and domain of sarcasm. This can lead to frail sarcasm detection models because different datasets come from different sources such as social media \cite{abu-farha-etal-2022-semeval, khodak_large_2018}, online forums \cite{oraby-etal-2016-creating}, TV series \cite{castro_towards_2019}, product reviews \cite{filatova2012irony}, or conversations \cite{chakrabarty-etal-2022-flute}. The vast range of domains, styles, and topics that sarcasm can be associated with invites the question of whether sarcasm detection models fine-tuned on a specific dataset are generalizable to other datasets. 

Also, existing datasets of sarcasm contain sarcasm labels from different sources: tags (e.g., \#sarcasm or /s) \cite{khodak_large_2018, joshi_harnessing_2015}, labels by annotators \cite{castro_towards_2019, oraby-etal-2016-creating, riloff_sarcasm_2013}, or labels by the authors of the posts  \cite{oprea_isarcasm_2020}. Collecting data with labels based on tags is easy and scalable, but the data tend to be noisy and it poses a high risk of including false positives in the dataset \cite{khodak_large_2018, sykora2020qualitative}. Most manually annotated data come with labels provided by crowd workers or experts (though it raises the question of ``who is an expert of sarcasm?"). These labels are more reliable than tags, but they still pose a risk of wrongly reflecting the intention of the original utterance maker \cite{oprea_isarcasm_2020} as sarcasm can be missed (referred to by some as \textit{sarchasm}) even in multimodal settings \cite{fox2020sarchasm}. In text-only settings, which is what a lot of sarcasm detection models are based on, the chance of misinterpreted sarcasm would naturally be much higher as the available cues are more limited. Recent work has reported that previous state-of-the-art models fine-tuned on datasets with third-party labels showed vastly different (poorer) performance when predicting sarcasm on data with author labels, going from an F-score of 0.75 to 0.33 \cite{abu-farha-etal-2022-semeval, oprea_isarcasm_2020}. 

Thus, in this work, we investigate the generalizability of sarcasm detection models fine-tuned on different datasets that contain sarcasm from different domains (social media/online vs.\ offline conversations/dialogues), styles (aggressive vs.\ humorous), and labels from different perspectives (author labels vs.\ third-party labels). 
Specifically, we make the following contributions:

\noindent \textbf{C1.} Through cross-dataset comparisons, we test the generalizability of sarcasm detection models, which have been fine-tuned on various datasets, to other datasets. We identify the datasets that lead to the highest prediction accuracy scores across datasets.

\noindent \textbf{C2.} We test the effect of label source (author vs.\ third-party) on model performance by using a new dataset that we release.

\noindent \textbf{C3.} We analyze different styles of sarcasm found in different datasets and argue that future work should include a more diverse shapes of sarcasm.

\section{Related work}

If there is any consensus around the topic of sarcasm, it is arguably the fact that sarcasm is hard to define \cite{fox2020sarchasm}. But much is known about the communicative functions of sarcasm as they have been extensively studied in theoretical and experimental work. Sarcasm can be used to mock \cite{gibbs_irony_2000, pexman_does_2002}, to criticize harsher \cite{colston1997salting}, to be humorous \cite{gibbs_irony_2000, glucksberg1995commentary, matthews_roles_2006}, or to save one's own face \cite{jorgensen_functions_1996}. Sarcasm is not only multifaceted in its functions but also manifests itself in diverse forms, including hyperbole, often accompanied by interjections and intensifiers \cite{joshi_automatic_2017-1}, rhetorical questions \cite{oraby-etal-2016-creating}, violation of maxim of relevance \cite{sperber1986relevance} or maxim of quality \cite{grice1975logic}. 

Despite the complications surrounding this topic, a lot of work in computational linguistics came up with operational definitions of sarcasm for automatic sarcasm detection (i.e., contextual incongruity \cite{joshi_harnessing_2015}, and discrepancy between positive sentiment and negative situation \cite{riloff_sarcasm_2013}). Sarcasm detection has since had fluctuating success rates from an F-score of 0.51 to 0.97 \citep{buaroiu2022automatic}.

One reason behind varying results may come from the nature of the data. Twitter is the source that most previous computational work on sarcasm has relied on as training data \cite{barbieri2014modelling, joshi_harnessing_2015, ptavcek2014sarcasm, van-hee-etal-2018-semeval}. Other sources of sarcasm data include online forums \cite{oraby-etal-2016-creating, walker-etal-2012-corpus}, Reddit \cite{khodak_large_2018}, or TV series \cite{castro_towards_2019}. Data sources matter for building a more robust sarcasm detector; \citet{joshi_harnessing_2015} showed that the same models performed quite differently across different datasets, with variations in F-scores of around 0.20.

More recently, the importance of addressing the sources of sarcasm labels has also been raised in multiple studies \cite{abu-farha-etal-2022-semeval, oprea_exploring_2019, oprea_isarcasm_2020}. They collectively showed that sarcasm detection models perform significantly worse when making predictions on data with labels provided by the authors of sarcastic utterances. This is a factor worth examining as most datasets were created by having human annotators rate the level of sarcasm from the observer's point-of-view \cite{castro_towards_2019, riloff_sarcasm_2013, van-hee-etal-2018-semeval}.

In our work we investigate the robustness of sarcasm detection models by taking into account the label source, data source, types, and styles of sarcasm.

\section{Datasets}

We take four existing datasets of sarcasm and use them in our experiment. We select these datasets because they each have different elements that facilitate cross-data comparison. In Section \ref{sec:existing-data}, we provide descriptions for each dataset, followed by the strengths and limitations of each that motivate the creation of a new dataset, which we describe in Section \ref{sec:csc-creation}. We provide examples from each dataset in Table \ref{tab:examples}.

\begin{table*}[h!]
    \caption{Two examples from each of the four datasets to illustrate the different styles of sarcasm.}
    \label{tab:examples}
    \centering
    \resizebox{\textwidth}{!}{
    \begin{tabular}{llc}
        \toprule
        \textbf{Dataset} & \textbf{Ex. No.} & \textbf{Examples from each dataset} \\
        \midrule
        CSC& 1& \makecell[cl]{\textbf{Context}: Steve gives you a watering can on your birthday while smiling at you with a strange expression. \\But you don't even have a single plant. \\\textbf{Response}: Maybe I will use it as an outside shower.} \\
        \cmidrule{2-3}
         & 2& \makecell[cl]{\textbf{Context}: You know that Steve does almost everything at the last minute. Today Steve tells you that he has to \\ write a joint proposal with colleagues from a different team by the end of this week, and complains that \\ things are moving so slowly because everybody in the other team is a procrastinator. He says, "gosh, it's so \\ frustrating that I'm the only one trying to get the stuff moving. Everyone in this team is too laid back!" \\\textbf{Response}: Oh yeah, and you're not laid-back at all. You are the emperor of procrastination, Steve! If you're  \\the most highly motivated person on that team, they really must be the most unmotivated team in human history.}\\
         \midrule
       MUStARD & 1& \makecell[cl]{\textbf{Context}: 'How do I look?', 'Could you be more specific?', "Can you tell I'm perspiring a little?" \\\textbf{Response}: No. The dark crescent-shaped patterns under your arms conceal it nicely. }\\
        \cmidrule{2-3}
         & 2& \makecell[cl]{\textbf{Context}: "So. I just thought the two of us should hang out for a bit. I mean, you know, we've never really talked." \\ \textbf{Response}: I guess you'd know that, being one of the two of us, though, right? } \\
         \midrule
       SC V2 & 1&  \makecell[cl]{Aaahhh, so just not accomplishing our goals and going home is not defeat, \\ there has to be paperwork for it to be a defeat. Gotcha.} \\
        \cmidrule{2-3}
        & 2& \makecell[cl]{Ever hear of artificial ensemination? Why is that heteros only think there is \\one way to produce children? I find hetero sex disturbing, and an unnatural lifestyle choice.} \\
        \midrule
        iSarcasmEval &1 &  \makecell[cl]{Imagine going to university for 4 years when you could just follow Elon Musk on Twitter for free.}\\
         \cmidrule{2-3}
         & 2&  \makecell[cl]{The control button just fell off my laptop. How symbolic for my life.}\\
          \bottomrule
    \end{tabular}}
\end{table*}

\subsection{Existing resources}
\label{sec:existing-data}
\textbf{MUStARD} (The Multimodal Sarcasm Dataset) is a dataset of sarcasm that \citet{castro_towards_2019} curated out of several TV shows such as \textit{Friends} and \textit{The Big Bang Theory}. The dataset comes with transcripts, audio, and video files of 690 scenes. Each scene consists of contextual utterances followed by a target utterance, half of which are sarcastic (N~=~345). Each scene was annotated by 2 annotators, who gave binary sarcasm labels to each video. The inter-rater agreement was a Kappa score of 0.23 for the majority of the videos and 0.59 for the remaining ones; a third annotator resolved the possible disagreement.   

\noindent\textbf{Strengths:} MUStARD is a multimodal dataset, and the annotation was done based on both contexts and responses. Despite its multimodal nature, there is work that used the dataset in text-only settings as well \cite{das2023paralleled, fuzzy-sarcasm-zhang}.

\noindent\textbf{Weaknesses:} The dataset is relatively small, the annotator agreement is low, and the sentences are scripted (as opposed to naturally-occurring).

\noindent\textbf{Sarcasm Corpus V2} \cite{oraby-etal-2016-creating} is a dataset extracted from the Internet Argument Corpus, containing posts from 3 different online debate sites \cite{abbott-etal-2016-internet, walker-etal-2012-corpus}. Sarcasm Corpus V2 contains a total of 9,386 written sentences, half of which are sarcastic. Each sentence was annotated by 9 annotators, who gave binary labels after being shown context posts with a following response. The inter-rater agreement was 0.80 (in percent agreement) among 9 annotators and 0.89 among the 3 best annotators.\footnote{The agreement is for one of the three subsets. The authors do not report the agreement score for the entire dataset.}

\noindent\textbf{Strengths:} Sarcasm Corpus V2 is large, and several annotators contributed to the annotation based on both contexts and responses. 

\noindent\textbf{Weaknesses:} The dataset itself does not provide contexts, and the source domain (online debate sites) can bias the data into a specific type of language use that is not commonly found in regular conversations.

\begin{figure*}[h!]
    \centering
    \includegraphics[width=0.97\textwidth]{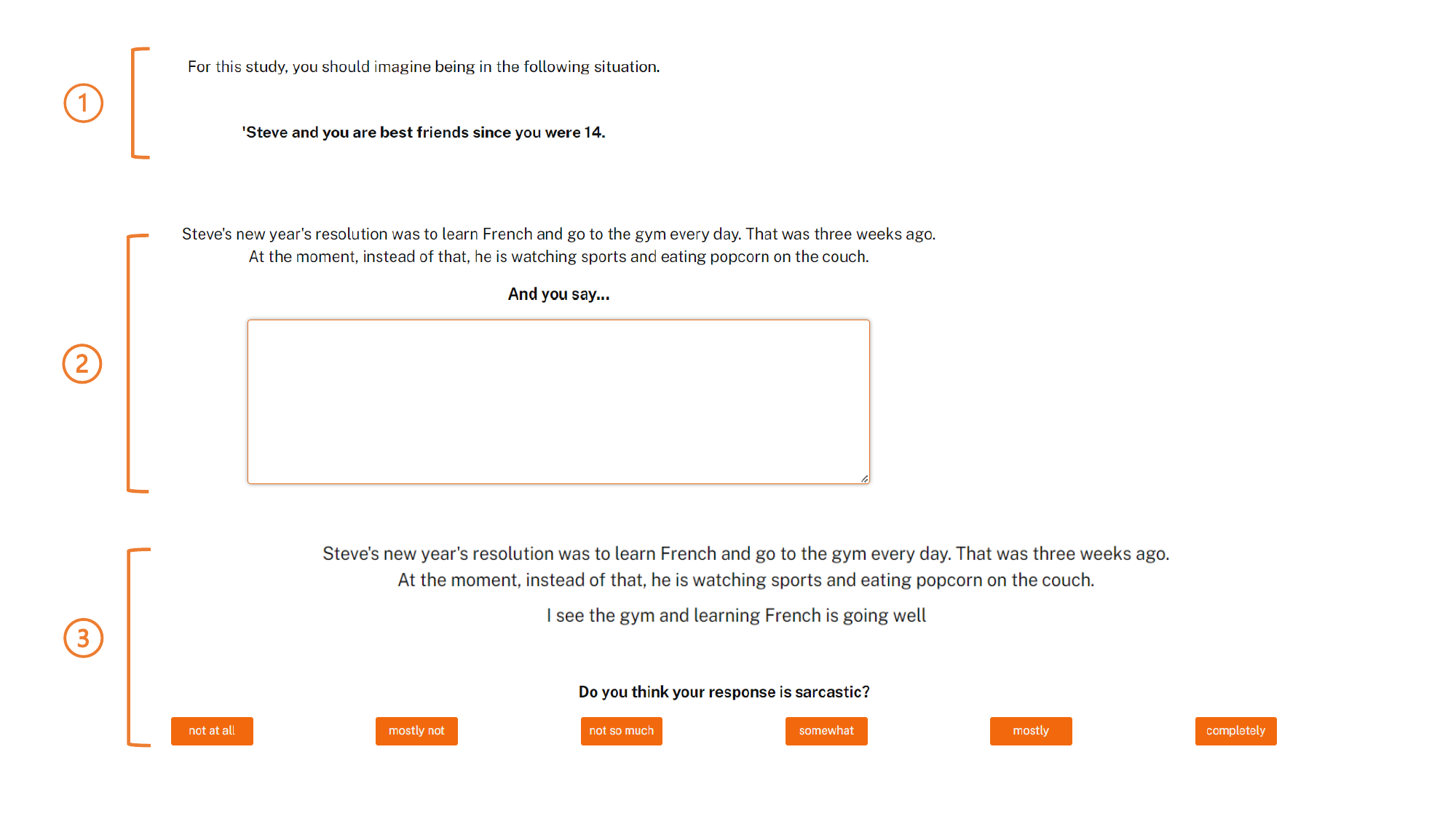}
    \caption{Data collection process illustration. Step 1 was shown to participants at the beginning of the experiment. All situations were shown as in Step 2 and afterwards, participants evaluated their own responses as shown in Step 3.}
    \label{fig:data-collection}
\end{figure*}

\noindent\textbf{iSarcasmEval} \citep{abu-farha-etal-2022-semeval} is a dataset containing Twitter posts and sarcasm labels by the authors of the posts. iSarcasmEval contains 3,801 sentences, 1,067 of which are sarcastic.\footnote{\citet{abu-farha-etal-2022-semeval} report slightly different numbers. The numbers reported in this paper are according to their data repository at \url{https://github.com/iabufarha/iSarcasmEval}.} Native English speakers on Prolific who were users of Twitter provided their own previous sarcastic and non-sarcastic posts 
with explanations of why they thought their posts were sarcastic. As the dataset has single author labels instead of third-party labels, no inter-rater agreement is applicable.

\noindent\textbf{Strengths:} iSarcasmEval provides author labels, which distinguishes itself from the previous datasets.

\noindent\textbf{Weaknesses:} It does not contain any contextual information.

\subsection{New dataset: Conversational Sarcasm Corpus (CSC)}
\label{sec:csc-creation}
We aimed to construct a new dataset that can overcome some of the main limitations of the existing datasets. To that end, we created a new sarcasm dataset that is based on offline conversational contexts (as opposed to Sarcasm Corpus V2 and iSarcasmEval), that is larger (than MUStARD and iSarcasmEval), and that contains naturally-occurring responses by a number of people (as opposed to MUStARD). 
The dataset contains author labels as well as third-party labels (as opposed to all the others). We constructed the dataset in two phases (part 1 and part 2) in order to first confirm our hypothesis of certain contextual prompts being more likely to trigger sarcastic responses. 
We then expanded our collection using the same approach but with different contextual prompts.

\begin{table*}[h!]
    \caption{Dataset comparison. A: authors labels, T:third-party labels. -C: without context, +C: with context, Sim.Conv.~=~Simulated conversations.}
    \label{tab:descriptive}
    \centering
    \resizebox{\textwidth}{!}{
    \begin{tabular}{lccccc}
        \toprule
      &  \textbf{CSC} & \textbf{MUStARD} & \textbf{SC V2} &  \textbf{iSarcasmEval}\\
        \midrule
        \# of sarcastic sentences & 2,210 (A) / 2,398 (T) & 345  &  4,693 & 1,067 \\
        \# of total sentences &  7,040 & 690  & 9,386 & 3,801\\
        \# of sent. for training & 4,420 (A) / 4,796 (T) & 690  &  9,386 & 2,134 \\
        Average sent. length (-C) & 11 & 12 & 49  & 18 \\
        Average sent. length (+C) & 51 & 42 & -  & - \\
        Author labels exist & Y & N  &  N & Y\\
        Third-party labels exist & Y & Y &   Y & N \\
        Is multimodal & N & Y  & N &  N\\
        Context exists &Y & Y  & N & N\\
        Original source domain & Sim. conversations & TV series &   Online debates & Social media  \\
        Original label type & Multi (1-6) & Binary  & Binary & Binary\\
        Annotator agreement & Moderate (Kendall's W 0.56) & Low (Kappa 0.23) & High (Percent agreement 0.80) & N/A\\
          \bottomrule
    \end{tabular}}
\end{table*}

In part 1\footnote{The data collection process of part 1 is published at \citet{jang2023intended} with analyses unrelated to this work. It provides more information about specific design choices of the stimuli.}, we created 32 situational descriptions, sometimes followed by an utterance by an interlocutor (The example in Figure \ref{fig:data-collection} ends with a description, but some stimuli were followed by an additional utterance by Steve). Half of the situational descriptions were created in a way in which the interlocutor was behaving in silly or annoying manners. We assumed that such situations would trigger more sarcastic responses than others based on our qualitative analysis of the MUStARD dataset. We recruited 60 (30 female, mean age~=~33) native English speakers online\footnote{We used Prolific and FindingFive for all our data collection processes.}. They were asked to imagine being a close friend of the interlocutor in the 32 situations (Step 1 in Figure \ref{fig:data-collection}). They freely responded to all situations by typing (Step 2 in Figure \ref{fig:data-collection}), and after this process was finished, they rated the level of sarcasm in their own responses (Step 3 in Figure \ref{fig:data-collection}) by selecting one of the following six options (1 - 6 Likert scale: \textit{not at all}, \textit{mostly not}, \textit{not so much}, \textit{somewhat}, \textit{mostly}, \textit{completely}). We collected ratings in a 6-point scale to obtain labels suitable for different
sarcasm modeling tasks (sarcasm classification with binary or multi-class labels and sarcasm quantification). As they were not asked to rate their responses until they freely responded to all situations, participants were not primed to be sarcastic, hence providing natural responses, some of which turned out to be sarcastic due to the influence of the provided context. 

Next, we recruited 360 evaluators (180 female, mean age~=~35) to rate the level of sarcasm of the responses given by each of the 60 authors of those responses. Six annotators saw the same 32 situations and the responses collected in the response generation process. Using the same 6-point scale, they rated the level of sarcasm of the last response in each situation. As there were 60 response generators in the previous process, the total number of annotators was 360 (6 * 60). 

The inter-annotator agreement indicated by Kendall's W was moderate (0.46). 

After successfully collecting ~30\% of sarcastic responses from part 1 (see Table \ref{tab:csc}), we collected more data points using the same experimental design but with different situational prompts. We created 40 new situation descriptions and collected responses from 128 native English speakers (64 female, mean age=37) and the sarcasm labels. This time, to increase variability in the generated sentences, participants were asked to imagine being a close friend of the interlocutor for half of the situations and a colleague for the other half. 
512 annotators (256 female, mean age=38) rated the level of sarcasm of the responses of each of the response generators. The Kendall's W was 0.56 indicating moderate agreement again.

\begin{table}[h!]
    \caption{Proportions of sarcastic (S) and non-sarcastic (NS) responses by author labels (A) and third-party labels (T) from the data collection process of CSC (Part 1 and Part 2). The original 1--6 labels were binary-coded for this table.}
    \label{tab:csc}
    \centering
    \resizebox{0.7\columnwidth}{!}{\begin{tabular}{ll rrrr}
        \toprule
         && \textbf{Part1} & \textbf{Part2} &  \textbf{Total} &  \textbf{\%}\\
         \midrule
         \multirow{2}{*}{A} & NS & 1,289 & 3,537 & 4,826 & 69  \\
           & S & 631 & 1,579  & 2,210 & 31 \\
         \cmidrule{1-6}
        \multirow{2}{*}{T} & NS & 1,307 &3,331 & 4,638 & 66 \\
        & S & 613 &1,785 & 2,398 & 34  \\ 
          \bottomrule
    \end{tabular}}
\end{table}

The CSC dataset thus has the following structure: context, response, response author id, evaluator id, author label for sarcasm, and third-party label for sarcasm. Out of 7,040 collected sentences, around 30\% were judged as sarcastic, which is a higher proportion of sarcasm than is usual (8 - 12\%), as reported in prior work \citep{oraby-etal-2016-creating, gibbs_irony_2000}. We release the full dataset\footnote{The full data can be found at: \\ \url{https://github.com/CoPsyN/CSC}.}, but we run all subsequent experiments with binary-coded labels (\textit{completely}, \textit{mostly}, \textit{somewhat} to \textit{sarcastic}) and with balanced number of labels after random downsampling (extracted using the random state of 2) to maximize comparability with the other datasets that are balanced.

\subsection{The content of the datasets}
\label{sec:data-comparison}
Comparing different datasets helps identify the vast spectrum of sarcasm available and further supports the need to test the generalizability of sarcasm detection models on the other datasets for a more robust sarcasm detector system. Table \ref{tab:descriptive} shows points of comparison among the 4 datasets. Apart from these quantitative and structural differences, here we report some qualitative differences (i.e., styles and topics) we find from inspecting each dataset. For CSC and MUStARD, with the contexts mostly being friendly situations, the elements of comedy and friendly mocking are prevalent. The contexts in MUStARD often do not provide enough information to indicate that the following utterance should be sarcastic or not, because more contextual information should be extracted either at the episodic or series level, or from multimodal cues, which often gets lost in the text-only version. Sarcasm Corpus V2 has the most aggressive, critical, and provocative type of sarcasm, which is understandable as the source of the data is online debate forums, from posts on controversial topics such as `homosexuality', `abortion', `gun control', or `religion'. In contrast, iSarcasmEval has the most instances of self-deprecating type of sarcasm since tweets are often used by  individuals to express their own opinions on a wide range of everyday topics. Even sarcastic instances that are directed at others in iSarcasmEval exude more distant/cynical attitudes towards the target compared to Sarcasm Corpus V2 (see Example No. 1 of iSarcasmEval in Table \ref{tab:examples}).

\section{Experimental setup}

\subsection{Models}
\label{sec:models}

We experimented with three encoder-only models from Transformers: BERT \cite{devlin-etal-2019-bert}, RoBERTa \cite{liuEtAl2019}, and DeBERTa \cite{deberta}. We limited our models to three widely-used and reliable classification models to focus on cross-data comparisons. 
We fine-tuned \texttt{bert-base-uncased}, \texttt{roberta-base}, and \texttt{mdeberta-v3-base}. The models were fine-tuned for 2 epochs with a batch size of 64, a learning rate of 5e-5, and a weight decay of 1e-2. We used 5-fold validation and 4 different seeds (1, 11, 21, 31). The fine-tuning was implemented using the Trainer class from the Hugging Face library, and conducted on a NVIDIA A100-PCIE GPU with a total memory of 40GB. All the results reported in Section \ref{sec:discussion} are an average across all seeds and folds.

\subsection{Intra-dataset vs.\ cross-dataset detection of sarcasm}
To test the generalizability of sarcasm detection spanning different datasets, we performed sarcasm detection in two ways: intra-dataset and cross-dataset detection. For the intra-dataset settings, all models fine-tuned on dataset A gave predictions on the hold-out test set of dataset A. For the cross-dataset settings, all models fine-tuned on Dataset A gave predictions on Dataset B. For all of these predictions, we did a 5-fold validation (64\% training, 16\% validation, 20\% test). 

\begin{table*}[h!]
    \caption{F-scores of all intra- and cross-dataset predictions. A: with author labels, T: with third-party labels, +CONT: text consisting of both context and utterance. The best fine-tuned LM(s) for each test set marked in bold (columnwise).}
    \label{tab:results}
    \centering
    \resizebox{\textwidth}{!}{\begin{tabular}{l l | c| c c  c c c c c c}
        \toprule
         \multirow{4}{*}{\textbf{Model}} & \multirow{4}{*}{\textbf{fine-tuned on}} & \multicolumn{9}{c}{\textbf{Predicted on}}  \\
            \cmidrule{3-11}
            & & \textbf{Intra-dataset} & \multicolumn{8}{c}{\textbf{Cross-dataset}} \\
        \cmidrule{3-11}
        \multicolumn{2}{c|}{} &  & \textbf{CSC-A+CONT} &\textbf{CSC-T+CONT} & \textbf{CSC-A} & \textbf{CSC-T} & \textbf{MUS+CONT} & \textbf{MUS} &  \textbf{SC V2} & \textbf{iSarcasm}\\
        \midrule
        \multirow{8}{*}{BERT} & CSC-A+CONT & 0.68& - & - & - & - & 0.54&0.56&0.42&0.50 \\
        & CSC-T+CONT &0.73 & - & - & - & - & 0.55&0.57&0.51&\textbf{0.53}\\
        & CSC-A & 0.67 & - & - &- & - & \textbf{0.57}&\textbf{0.58}&0.39&0.43 \\
        & CSC-T & 0.70& -&-&-&-& 0.56&0.56&0.46&0.47\\
        & MUS+CONT & 0.63 & 0.45 & 0.46 & 0.51 & 0.50 &- & - &  0.39& 0.44 \\
        & MUS &0.63& 0.47& 0.47& \textbf{0.53}& \textbf{0.52}& - & - & 0.40 & 0.45\\
        & SC V2 & \textbf{0.77} & 0.44&0.44&0.44&0.44& 0.39& 0.46&- & 0.45\\
        & iSarcasm &0.59 &\textbf{0.48}&\textbf{0.48}&0.52&0.51&0.44&0.50&\textbf{0.59}& -\\
        \midrule
        \multirow{8}{*}{RoBERTa} & CSC-A+CONT &0.68 & - & - &- &- & \textbf{0.59}& 0.55&  0.48& 0.52\\
        & CSC-T+CONT &0.72& - & - & -& -& 0.57 & \textbf{0.57}&\textbf{0.56} & \textbf{0.54}\\
        & CSC-A & 0.66 & -&-&-&-& 0.55& 0.56&0.42 & 0.44\\ 
        & CSC-T & 0.70 &-&-&-&-& 0.56& \textbf{0.57}& 0.51&0.51 \\
        & MUS+CONT &0.44& 0.35& 0.35 & 0.39 & 0.38 &-&-&  0.37 & 0.38\\
        & MUS &0.44 & 0.35& 0.35& 0.41 & 0.40 & - & - & 0.36 & 0.40\\
        & SC V2 &\textbf{0.80}& \textbf{0.47}& \textbf{0.49}& \textbf{0.52}& \textbf{0.53}& 0.39& 0.49&  -& \textbf{0.54}\\
        & iSarcasm &0.42& 0.36& 0.35& 0.38& 0.39& 0.36& 0.37&  0.44&- \\
        \midrule
        \multirow{8}{*}{DeBERTa} & CSC-A+CONT & 0.67& - & - & - & - & \textbf{0.55}&\textbf{0.57}&0.44&\textbf{0.52} \\
        & CSC-T+CONT &0.72 & - & - & - & - & \textbf{0.55}&0.56&0.53&\textbf{0.52}\\
        & CSC-A & 0.65 & - & - &- & - & 0.54&0.55&\textbf{0.56}&0.48 \\
        & CSC-T & 0.69& -&-&-&-& 0.53&0.54&0.55&0.50\\
        & MUS+CONT & 0.44 & 0.37 & 0.37 & 0.40 & 0.40 &- & - &  0.45& 0.39 \\
        & MUS &0.43& 0.35& 0.35& 0.43& 0.41& - & - & 0.36 & 0.40\\
        & SC V2 & \textbf{0.78} & \textbf{0.53}&\textbf{0.53}&\textbf{0.50}&\textbf{0.50}& 0.37& 0.47&- & 0.49\\
        & iSarcasm &0.41 &0.34&0.34&0.38&0.37&0.45&0.50&0.35& -\\
        \bottomrule
    \end{tabular}}
\end{table*}

\section{Results and discussion}
\label{sec:discussion}

Table \ref{tab:results} reports the F-scores of all models in the intra-dataset and cross-dataset conditions. 

\subsection{Intra-dataset predictions}

For all LMs, the best performance is obtained for the Sarcasm Corpus V2 (SC V2), followed by the Conversation Sarcasm Corpus with third-party labels (CSC-T), and the lowest performance on iSarcasmEval. The high performance of SC V2, aside from it being the largest dataset with the most annotators, is also attributed to its source: online forums. In these forums, users can only use text for communication, potentially leading to a rich concentration of lexical cues associated with sarcasm, such as aggression and negative emotions. These textual cues are more easily identifiable by LMs, enhancing their ability to detect sarcasm. MUStARD fares worse than most other models possibly due to the loss of multimodal cues in text-only settings, which normally would have complemented the long-dependency contexts.

\noindent \textbf{The source of sarcasm labels} consistently affects the model performance in the intra-dataset settings. Models fine-tuned on iSarcasmEval, which only has author labels, show the worst performance compared to the others. For CSC also, LMs fine-tuned with third-party labels always perform better than with author labels (\textbf{C2}).  This observation could suggest that language models may act more as passive observers lacking introspective abilities: Observers, either humans or language models, must rely on external cues and make inferences about an utterance to interpret it since they have no direct access to the complex motivations of the speaker. This aligns with the existence of \textit{sarchasm}, `sarcasm gone missed', which is a prevalent phenomenon in human communication \cite{fox2020sarchasm}.

\subsection{Cross-dataset predictions}

\textbf{Overall}, all LMs struggle to detect sarcasm on the other datasets proportionately to their performance in intra-dataset settings. Even LMs fine-tuned on SC V2, which showed the highest performance in the intra-dataset predictions, do not generalize nearly as well to the other datasets. 

The datasets that result in the most struggle for fine-tuned LMs in generalization are SC V2 (BERT), MUStARD (RoBERTa, DeBERTa), and iSarcasmEval (RoBERTa), based on the average score of each row in Table \ref{tab:results}. In contrast, different versions of CSC used in fine-tuning lead to the best performance in stable generalization to the other datasets for all LMs, except for one case (BERT+iSarcasmEval performed the best on SC V2 with an F-score of 0.59). 

Findings from previous work employing similar methodologies suggest that, for models to generalize to other datasets effectively, they must first exhibit robust performance on the data used for their fine-tuning \cite{FORTUNA2021102524} or that datasets should be large \cite{yin2021towards, halevy2009unreasonable}. But, our results show that the new CSC dataset can deal with a broader range of sarcasm despite not being the largest dataset nor yielding the highest accuracy in predictions on its own dataset. Such relative competence could have stemmed from a strong advantage of CSC over other datasets, which is the psycholinguistically-motivated collection and annotation of the data. This suggests that an additional factor in a dataset that contributes to a high generalizability of an LM includes data collection methodology.

\noindent \textbf{The effects of context} were tested by fine-tuning the LMs with or without context for CSC and MUStARD. For the \textit{with-context} condition, contexts were concatenated with responses for fine-tuning; for the \textit{without-context} condition, only responses were used in fine-tuning. This was done because context is an important aspect in humans' processing of sarcasm but it has been less obvious for LMs \cite{woodland_context_2011, castro_towards_2019, jaiswal2020neural, ghosh_2018_context}. Also, testing the effects of context was important for CSC as it has an embedded structure in which different responses are preceded by the same context. 

The results showed that LMs fine-tuned with context obtain slightly better results than LMs fine-tuned without it (in intra-dataset predictions), though the improvement was very small (between 0.01 and 0.03). In cross-dataset predictions, the presence of context did not affect the generalizability of the models either. Note that this could be the result of the way context was concatenated to the response or of the number of fine-tuning epochs. For more comprehensive results, future research should address this aspect with targeted experimental manipulations.

\noindent \textbf{The domain of the dataset} does not affect the model performance in any obvious way. One might assume that the generalization ability of LMs would benefit from datasets that share a domain (i.e., CSC and MUStARD; both from conversational contexts). This was not the case, as only the LMs fine-tuned on CSC predicted sarcasm on MUStARD well (F-scores 0.53 - 0.59), but not the other way around (F-scores 0.35 - 0.53) except for two cases (BERT+MUS on CSC-A and CSC-T). 

The LMs that generalized the best on CSC were in fact fine-tuned on SC V2 (for RoBERTa and DeBERTa) or iSarcasmEval or MUStARD (for BERT), most of which come from a different domain than CSC (social media or debate forums). The success of SC V2 may be attributed to the size and annotation quality of the dataset \citep{yin2021towards}.

\vspace{0.1in}
\noindent \textbf{Combining all datasets for fine-tuning} did not improve model performance. When BERT was fine-tuned on all datasets combined, the average F-score was 0.71, still lower than the two highest performing datasets (SC V2 and CSC-T-CONT). This shows that a mere combination of several datasets with different sizes, styles, and label sources is insufficient in improving the generalizability of sarcasm detection.

\section{Post-hoc qualitative analysis}
\label{sec:posthoc}

We probe for reasons behind the low generalizability of LMs with a post-hoc analysis. 
We conducted a quantitative analysis in which we quantitatively analyze and demonstrate how LMs get accustomed to different types of cues for detecting sarcasm when fine-tuned with different datasets (\textbf{C3}). We focus on BERT only for this analysis.

\begin{table}[h!]
\centering
    \caption{List of LIWC categories and examples used for Section \ref{sec:posthoc}. Original source at:  \url{https://mcrc.journalism.wisc.edu/files/2018/04/Manual_LIWC.pdf} }
    \label{tab:liwc-categories}
    \centering
    \resizebox{\columnwidth}{!}{\begin{tabular}{ll}
        \toprule
      \textbf{Category} & \textbf{Examples} \\
      \hline
     Negations& no, not, never \\
       Positive emotion& love, nice, sweet\\
      Negative emotion& hurt, ugly, nasty \\
      Anxiety& worried, fearful \\
      Anger& hate, kill, annoyed\\
      Sadness& crying, grief, sad\\
      Social processes & mate, talk, they \\
      Family& daughter, dad, aunt \\
      Friend& buddy, neighbor\\
      Cognitive processes& \makecell[tl]{cause, know, ought, \\think, know, because, \\ should, maybe, always, never} \\
      Perceptual processes & look, heard, feeling\\
       Drives & \makecell[tl]{friend, social, win, success, \\ superior, bully, take, \\ prize, benefit, danger, doubt}\\
      Religion& altar, church\\
      Swear words& fuck, damn, shit\\
      Online register & btw, lol, thx\\
      Agreement & agree, ok, yes\\
      Nonfluencies& er, hm, umm\\
          \bottomrule
    \end{tabular}}
\end{table}

\noindent\textbf{Methods.} We identified certain patterns of language found in each dataset that may have led to differing results. 
As a proxy for \textit{styles} or \textit{registers}, we chose 17 semantic and psycholinguistic categories provided by the linguistic analysis tool LIWC \cite{pennebaker}, which quantitatively analyzes text in terms of psychological constructs such as emotion, cognition, or perception, among others (See Table \ref{tab:liwc-categories}). LIWC produces the ratio of words belonging to each of these categories per sentence.

Working with one dataset at a time, we took all the sarcastic instances 
correctly identified as sarcastic 1) \textit{only} by BERT fine-tuned on the same dataset (\textit{own success}) and  2) \textit{only} by BERT fine-tuned on the other datasets (\textit{others' success}). For each of these instances, we obtained LIWC scores for the 17 linguistic categories and averaged them across all the instances. We calculated the difference between the scores of \textit{own success} and \textit{others' success}. The score difference indicates the linguistic properties that facilitate the detection of sarcasm that are uniquely present in one dataset (and not in the others).

\begin{figure}[h!]
    \centering
    \includegraphics[width=\columnwidth]{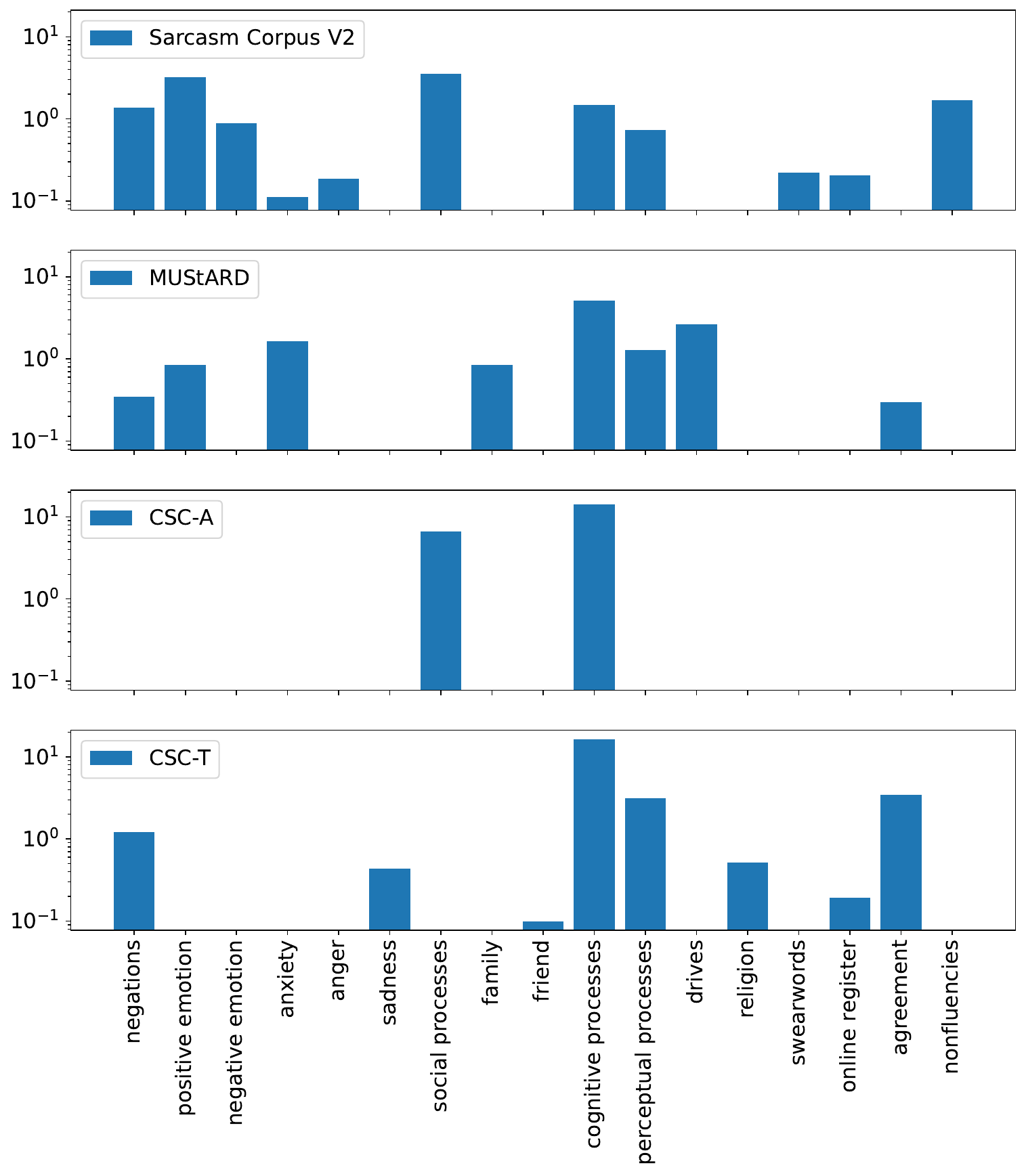}
    \caption{Post-hoc analysis to quantify diverging features found in different datasets of sarcasm. The height of each bar represents the log transformed difference between the score for each linguistic property found in sentences uniquely detected by LMs fine-tuned on the same data and those uniquely detected by LMs fine-tuned on the other data. The full description of the categories is provided in Table \ref{tab:liwc-categories}.}
    \label{fig:posthoc_diff}
\end{figure}

\noindent\textbf{Results.} 
Figure \ref{fig:posthoc_diff} reports the unique features that help BERT detect sarcasm in each dataset. A major characteristic of \textbf{Sarcasm Corpus V2} is that it contains many words related to negative emotions such as anger, and words related to social issues, swearwords, online-style words, and disfluencies. This matches our preliminary description of the dataset described in Section \ref{sec:data-comparison} that Sarcasm Corpus V2 contains the most aggressive and critical types of sarcasm of out of all the datasets. \textbf{MUStARD} contains a lot of words related to family and drives (i.e., achievement, risk, reward, etc.). We suspect that this phenomenon may be influenced by the TV series The Big Bang Theory, where the characters often converse on topics related to achievement. \textbf{CSC} shows a lot of words related to agreement (i.e., Ok, yes, etc.) possibly because the speakers were instructed to talk to a close friend or colleague. The analysis also reveals a significant presence of words associated with religion, predominantly due to the use of terms like `god' or `Jesus' in expressions such as `oh my god' or `Jesus Christ'. This trend might be attributed to numerous situational prompts where the interlocutor's behavior is portrayed as silly, eliciting strong emotional responses. The results for \textbf{iSarcasmEval} are not reported because there were no instances for which fine-tuning on iSarcasmEval only and not on the other datasets produced correct predictions. 

The post-hoc analysis of different datasets and model performance shows that sarcasm does indeed come in different shapes and domains, some serious and genuinely meant to inflict verbal pain to strangers, and some humorous and friendly used among well-meaning acquaintances, which is also supported by prior psycholinguistic work \cite{bowes_when_2011, colston1997salting, dews_why_1995, frenda2022unbearable, gibbs_irony_2000, pexman_does_2002}.

\section{Conclusion}

We conducted intra- and cross-dataset sarcasm detection experiments to examine the robustness of sarcasm detection models. We introduced a new dataset, which we compared with other existing datasets of varying characteristics. Several characteristics regarding sarcasm in each dataset, such as label source (authors vs.\ third-party), domain (social media/online vs.\ offline conversations/dialogues), style (aggression vs.\ harmless mocking) were used as points of comparison. All LMs gave better predictions on the same dataset they were fine-tuned on and showed much lower prediction performance on the other datasets (\textbf{C1}). Still, LMs fine-tuned on our new dataset CSC generalized the best to the other datasets. This was the case even when the domain of sarcasm in the target dataset was different from that of CSC. LMs performed better when the ground-truth labels were from third-party annotators rather than authors themselves (\textbf{C2}). A post-hoc analysis and a closer look at each dataset supported our assumption that such low cross-dataset predictions may be attributed to sarcasm coming in various styles, intent, and shapes (\textbf{C3}). Therefore, future research should take the vast scope of sarcasm into consideration rather than only focusing on a narrow definition of sarcasm as `the utterance of the opposite of the intended meaning' or as `a figure of speech intended to be hurtful, insensitive, and critical'.

\section*{Limitations}

We limited the number of datasets and language models to focus on the performance comparison of language models fine-tuned on different datasets. Other publicly available sarcasm datasets not used in this work include SAD \cite{painter-etal-2022-utilizing}, SARC \cite{ptavcek2014sarcasm}, and MUStARD++ \cite{ray-etal-2022-multimodal}. The language models mentioned in this paper exclude generative models or dialogue-focused models. Experiments employing other models with more parameters, other datasets, and with different hyperparameters (i.e., different number of epochs, etc.) are left to future work.

\section*{Ethics Statement}

We generally see little ethical issue related to this work. All our experiments involving human participants were conducted on a voluntary basis with a fair compensation suggested by the recruitment platform Prolific and participants were informed on how the data will be used. All our modeling experiments were conducted with open-source libraries, which received due citations. The experiment is in line with the ethical regulations of the University of Konstanz (IRB number 05/2021) . But we acknowledge that at times, sarcasm in itself can be a sensitive topic including offensive language and content depending on the circumstances.

\section*{Acknowledgements}
We thank Bettina Braun and Hsun-Hui Lin for their contribution in the data collection process. We thank Sabastian Padó and Sabine Schulte im Walde for their advice on the initial draft of this paper. We also thank the anonymous reviewers for their feedback.

\bibliographystyle{acl_natbib}
\bibliography{acl2023, anthology}

\end{document}